%% file: main.tex
\pdfoutput=1
\documentclass[11pt]{article}

\usepackage{ACL2023}

\input{preamble}

\title{Soft Token Attacks Cannot Reliably Audit Unlearning in Large Language Models}

\author{Haokun Chen\thanks{Work done while at Intel Labs.} \\
  Ludwig Maximilian University of Munich \\ Munich Center for Machine Learning (MCML)\\
  \texttt{haokun.chen@campus.lmu.de} \\\And
  Sebastian Szyller\footnotemark[1] \\
  Aalto University\\
  \texttt{contact@sebszyller.com} \\\AND
  Weilin Xu \\
  Intel Labs\\  
  \texttt{weilin.xu@intel.com} \\\And
  Nageen Himayat \\
  Intel Labs\\
  \texttt{nageen.himayat@intel.com}
  }

\begin{document}
\maketitle

\input{0abstract}
\input{1intro}

\input{2background}
\input{3method}
\input{4eval}
\input{5discussion}
\input{6acknowledgments}
\bibliography{main}
\input{appendix}

\end{document}

%% file: preamble.tex
\usepackage{times}
\usepackage{latexsym}
\usepackage{hyperref}

\usepackage[T1]{fontenc}
\usepackage[utf8]{inputenc}

\usepackage{microtype}

\usepackage{inconsolata}

\usepackage[toc,page]{appendix}
\usepackage{amsmath,amssymb,amsfonts}
\usepackage{color}
\usepackage[inline]{enumitem}
\usepackage{graphicx}
\usepackage{multicol}
\usepackage{multirow}
\usepackage{textcomp}
\usepackage{xcolor}
\usepackage{xspace}
\usepackage{tabularx}
\usepackage{algorithm}
\usepackage{algpseudocode}
\usepackage{amsmath}
\usepackage{algorithmicx}
\usepackage{subcaption}

\newcolumntype{Y}{>{\centering\arraybackslash}X}

\newcommand{\fauxsection}[1]{\noindent\textbf{#1}\xspace}
\newcommand{\tofu}{\textsl{TOFU}\xspace}
\newcommand{\whp}{\textsl{WHP}\xspace}
\newcommand{\sta}{\textsl{STA}\xspace}

%% file: 0abstract.tex
\begin{abstract}
% LLMs are popular
% Privacy, IP, and harmful output considerations
% Unlearning is one of the key reserach angles to address both
% Is approximate unlearning succesful
% You can attack/audit with soft-prompts
% Soft-prompts are too strong and prior eval cannot be taken for granted
Large language models (LLMs) are trained using massive datasets.
However, these datasets often contain undesirable content, e.g., harmful texts, personal information, and copyrighted material.
To address this, \emph{machine unlearning} aims to remove information from trained models.
Recent work has shown that soft token attacks (\sta) can successfully extract unlearned information from LLMs.
In this work, we show that \sta{s} can be an inadequate tool for auditing unlearning.
Using common unlearning benchmarks, i.e., \textit{Who Is Harry Potter?} and \textit{TOFU}, we demonstrate that, in a \emph{strong auditor} setting, such attacks can elicit any information from the LLM, regardless of (1) the deployed unlearning algorithm, and (2) whether the queried content was originally present in the training corpus.
Also, we show that \sta with just a few soft tokens ($1-10$) can elicit random strings over 400-characters long.
Thus showing that \sta{s} must be used carefully to effectively audit unlearning.
Example code can be found at \href{https://github.com/IntelLabs/LLMart/tree/main/examples/unlearning}{https://github.com/IntelLabs/LLMart/tree/main\\/examples/unlearning}
\end{abstract}

%% file: 1intro.tex
\section{Introduction}\label{sec:introduction}
% LLMs are popular
% XYZ considerations
% Unlearning is one of key reserach angles
% Is approximate unlearning succesful
% You can audit with soft-prompts
% We show that auditing unlearning with soft prompts is too strong
% Contribution:
% benchmarks with datasets and on llms
% --- point out that harry potter is broken in general because chat and no \n\n
% eliciting randoms strings is possible and invalidates regular unlearning benchmarks
% moving forward, researchers need to calibrate they solutions better and get better baselines
Large language models (LLMs) excel in many downstream tasks, e.g., machine translation~\cite{zhu2023multilingual}, content generation~\cite{acharya2023llm}, and complex problem-solving~\cite{chen2024dat}.
Their performance is attributed to their large-scale architectures that require datasets consisting of up to trillions of tokens to train effectively~\cite{kaplan2020scalinglaws}.
These datasets are typically derived from large-scale corpora sourced from public internet text.
However, such datasets can contain undesirable content, e.g., instructions for building weapons, violent or explicit material, private information, or copyrighted content.
Given the sensitive nature of such data, it may be necessary to remove it from the LLM to comply with the local regulations, or internal company policies.

\emph{Machine unlearning} is a tool for removing information from models~\cite{cao2015towards, bourtoule2021unlearning}.
\emph{Approximate unlearning} usually refers to removing information from models without retraining them from scratch~\cite{zhang2024right, eldan2023harrypotter, izzo2021approximate}, ensuring that the resulting model deviates from a fully retrained version within a bounded error.
While numerous studies have proposed various unlearning algorithms, most lack formal guarantees.
Prior research has demonstrated that many unlearning techniques can be circumvented through rephrasing of the original data~\cite{shi2024muse}.
Recent work has shown that a \emph{soft token attack} (\sta) can be used to elicit harmful completions and extract supposedly unlearned information from models~\cite{schwinn2024soft,zou2024circuitbreakers}.

In this work, we introduce a simple framework for \emph{auditing unlearning}, and demonstrate that \sta{s} are inappropriate for verifying the effectiveness of approximate unlearning in a \emph{strong auditor} setting.
We show that the auditor can elicit any information from the model, regardless of its training data. 
We claim the following contributions:
\begin{enumerate}[topsep=0pt, itemsep=0pt, parsep=0pt, label=\arabic*.]
    \item We show that \sta{s} effectively elicit unlearned information in all tested unlearning methods and benchmark datasets (\emph{Who Is Harry Potter?}, and \emph{TOFU}). Additionally, we show that \sta{s} also elicit information in the base models that \textbf{were not fine-tuned on the benchmark datasets} (Section~\ref{sec:eval:attacks}).
    \item We further demonstrate that the \sta{s} are inappropriate for evaluating unlearning -- we show that a single soft token can elicit $150$ random tokens, and ten soft tokens can elicit over $400$ random tokens (Section~\ref{sec:eval:random-strings}).
\end{enumerate}

% The remainder of this paper is organized as follows:
% In Section~\ref{sec:background} we provide an overview of the necessary background, and the related work.
% Section~\ref{sec:method} introduces a general auditing framework, and instantiates it using \sta.
% In Section~\ref{sec:eval}, we demonstrate the efficacy of \sta, and subsequently its failure, as a tool for auditing unlearning in LLMs. 
% In Section~\ref{sec:discussion} we discuss additional considerations for auditing unlearning in LLMs.
% We conclude the paper in Section~\ref{sec:conclusion}, and highlight some limitations in Section~\ref{sec:limitations}.

%% file: 2background.tex
\section{Background}\label{sec:background}
% \fauxsection{Large language models} (LLMs) process input text through an auto-regressive framework.
% Given an input sequence of tokens $x_{1:t}$, the model computes the conditional probability distribution $p(x_{t+1}|x_{1:t})$ over the vocabulary at each time-step.
% The likelihood of the sequence is given by:
% \begin{equation}
%     \log p(x_{t+1}|x_{1:t}) = \sum_{t=1}^T \log p(x_{t}|x_{1:t-1})
% \end{equation}
% At inference time, the tokens is generated iteratively by sampling the next token $x_{t+1}$ from this distribution (e.g., via greedy decoding or nucleus sampling~\cite{holtzman2019curious}), then appending it to the context $x_{1:t}$ for the subsequent step.
% \fauxsection{Soft \& Hard token attacks}: We formalize both hard and soft token attacks as follows. Let $T$ denote a tokenized input sequence consisting of $n$ tokens. \textit{Hard} token attacks aim to optimize or select multiple discrete token IDs, which are appended to the input sequence to elicit specific model behaviors—such as generating an affirmative prefix in the output. In contrast, \textit{soft} token attacks operate in the embedding space. The tokenized input $T$ is first projected by an embedding layer, which maps discrete token IDs into continuous embeddings of dimension $D$, i.e., $\mathbb{Z}^n \rightarrow \mathbb{R}^{n \times D}$. Soft token attacks, such as those proposed by \cite{schwinn2024soft}, optimize these embeddings directly to minimize a predefined objective function.
\fauxsection{Adversarial prompt} $x_a$ is an input prompt to the LLM, obtained by applying the transform $T(\cdot)$ to the base prompt $x_p$: $x_a = T(x_p, \textsl{aux})$ to elicit a desired completion $c$.
$T$ can be any function that swaps, removes or adds tokens; \textsl{aux} denotes any additional needed information.
Such arbitrary attacks are expensive to optimize, and difficult to reason about.
In practice, $T$ optimizes an \emph{adversarial suffix} $x_s$ that is appended to $x_p$ to elicit $c$~\cite{zou2023gcg}.
These suffix-only attacks also allow efficient use of the KV-cache~\cite{pope2023efficiently}.
Specifically, we optimize the probability:
\begin{equation}
    Prob = P(c | x_p \oplus x_s).
\end{equation}

\begin{figure}[t!]
    \centering
    \includegraphics[width=0.45\textwidth]{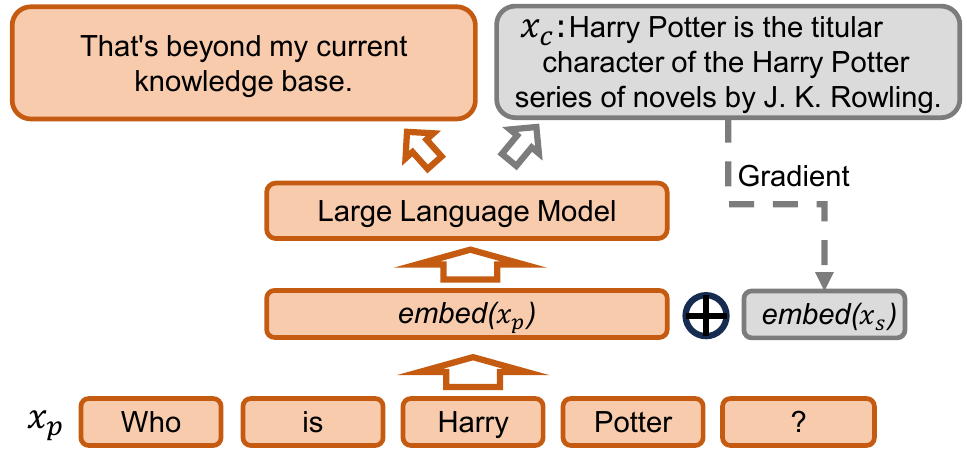}
    \vspace{-8pt}
    \caption{\sta combines $x_p$ with the optimized $x_s$.}
    \vspace{-12pt}
    \label{fig:attack}
\end{figure}

An adversary with white-box access to the LLM, can instead mount the attack in the \emph{embedding space} i.e. modify the \emph{soft tokens}:
\begin{equation}
    Prob = P(c | embed(x_p) \oplus embed(x_s)).
\end{equation}
In this case, $T$ uses the gradient from the LLM to update $x_s$.
We visualize such attack in Figure~\ref{fig:attack}.

\fauxsection{Machine unlearning} (MU) aims to remove information from models.
Consider a machine learning model $f$ trained using a training dataset $D_{train}$. 
During an unlearning request to remove a specified subset $D_{forget} \in D_{train}$, 
the objective of MU is to produce an unlearned model $f_{u}$ that eliminates the influence of $D_{forget}$.
There are two types of MU -- exact, and approximate unlearning.

\fauxsection{Exact unlearning} ensures the output distribution of $f_u$ is statistically indistinguishable from $f_{ret}$ -- a model retrained exclusively on the retained dataset $D_{retain} = D_{train} / D_{forget}$.
This guarantees provable data removal, satisfying:
\begin{equation}
\centering
\begin{aligned}
    p(f_u(x) = y) & = p(f_{ret}(x) = y) \quad \\ 
    & s.t. \quad \forall(x,y) \in D_{train}.
\end{aligned}
\end{equation}

It can be made more efficient by splitting the $D_{train}$ into overlapping chunks, and training an ensemble~\cite{bourtoule2021machine}.
During an unlearning request, only the models containing the requested records are retrained.
For certain classes of models, exact unlearning without retraining is possible, e.g. ECO adapts the Cauwenberghs and Poggio (CP) algorithm for exact unlearning within LeNet~\cite{huang2024eco}, and MUSE relabels the target data to achieve unlearning for over-parameterized linear models~\cite{yang2024muso}.

\fauxsection{Approximate unlearning} relaxes the strict equivalence requirement, it only requires that $f_{u}$ approximates $f_{ret}$ within some bounded error.
It relies on empirical metrics or probabilistic frameworks.
In LLMs, approximate unlearning is typically accomplished by overwriting the information in the model~\cite{eldan2023harrypotter,wang2024rkld}, guiding the model away from it~\cite{feng2024fine}, or editing the weights and/or activations~\cite{liu2024large,bhaila2024soft,li2024wmdp,tamirisa2024toward,huu2024effects, ashuach2024revs, meng2022locating, meng2022mass}.

\section{Related work}\label{sec:related}
While advances have been made in developing machine unlearning algorithms for LLMs, rigorous methodologies for auditing the efficacy of unlearning remain understudied.
Adversarial soft token attacks (\sta{s})~\cite{schwinn2024soft} and 5-shot in-context prompting~\cite{doshi2024doesunlearningtrulyunlearn} have been shown to recover unlearned knowledge in models. 
When model weights can be modified, techniques such as model quantization~\cite{zhang2024does} and retraining on a partially unlearned dataset ~\cite{lucki2024adversarial,hu2024jogging,chen2025does} have also proven effective in recalling forgotten information. 
~\cite{lynch2024eight} examined eight methods for evaluating LLM unlearning techniques and found that their latent representations remained similar.
News and book datasets are used to analyze unlearning algorithms from six different perspectives~\cite{shi2024muse}.
It was shown that fine-tuning on \emph{unrelated} data can restore information unlearned from the LLM~\cite{qi2024unrelateddata}, indicating the existing unlearning methods do not actually remove the information but learn a \emph{refusal filter} instead.
Several benchmarks have been developed to evaluate the existing unlearning algorithms.
Besides, an unlearning benchmark was introduced based on fictitious author information~\cite{maini2024tofu}.
For real-world knowledge unlearning, \emph{Real-World Knowledge Unlearning} (RWKU) used 200 famous people as unlearning targets~\cite{jin2024rwku}, while WDMP focused on unlearning hazardous knowledge in biosecurity, cybersecurity~\cite{li2024wmdp}.

%% file: 3method.tex
\section{Auditing with Soft Token Attacks}\label{sec:method:framework}
% -- Outline
% --- A framework for auditing approximate unlearning -- some algo?
% --- Instantiation of the framework using soft token attacks
An \emph{oracle} auditor $A_o$ takes an unlearned model $f_u$ and the candidate sentences $x_c \in X_c$, and outputs a \emph{ground} truth, binary decision $a=\{0, 1\}$ indicating whether the given records was part of $D_{train}$ of:
\begin{equation}
    a = A_o(f_u, X_c = D_{forget}, \textsl{aux}).
\end{equation}
$A_o$ is unrealistic but it can be easily instantiated for exact unlearning where $A_o$ knows the training data associated with $f$: $aux = \{  D_{retain} \}$.

A realistic unlearning auditor $A_u$ takes an $f_u$, and $D_{forget}$ and outputs a score $s=(0,1)$ indicating whether the records were in $D_{train}$:
\begin{equation}
    s = A_u(f_u, X_c = D_{forget}, \textsl{aux}=\emptyset)
\end{equation}
$A_u$ represents cases where users unlearn facts from models that they did not create, e.g. to prevent harmful outputs.

In this work, we instantiate the soft token attack auditor $A_\sta$ based on the soft token attacks (\sta{s}) against unlearning~\cite{schwinn2024soft,zou2024circuitbreakers}.
$A_\sta$ compares the relative difficulty of eliciting $c$ for $f_{ft}$ and $f_u$.
The unlearning procedure is effective if eliciting completions using $f_u$ is more difficult than $f_{ft}$.
\begin{equation}
    s = A_\sta(f_u, X_c = D_{forget}, \textsl{aux}=\{f_{ft}\}).
\end{equation}
Crucially,~\citealp{schwinn2024soft} optimized $x_s$ only w.r.t. the affirmative beginning of $x_c$ ($x_p=$`\emph{Who is Harry Potter?}', $x_c=$`\emph{Harry Potter is}').
Contrary to that minimal setting, our \emph{strong auditor} $A_\sta$ optimizes $x_s$ w.r.t. the entire $x_c$. 
Our goal is to study the extreme scenario where the auditor expects the exact completions for, e.g. to check for copyrighted content, or personal information.

In the Appendix, Figure~\ref{fig:auditor} gives an overview of the auditing procedure, and the difference between $A_o$ and $A_\sta$; Table~\ref{tab:notation} summarizes the notation.

%% file: 4eval.tex
\section{Evaluation}\label{sec:eval}
% HP and TOFU baselines
% Complete failure of said framework
% Evaluating with soft prompts isn't even possible -> random string experiments
% We can do it for the trained, unlearned, and even base.
% Btw, also random strings.

\subsection{Experiment setup}

\fauxsection{Datasets.}
For evaluation, we use two popular benchmark datasets:
\begin{enumerate*}[label=\textbf{\arabic*})]
    \item \textit{Who Is Harry Potter?}~\cite{eldan2023harrypotter} (\whp)
    that intends to remove formation about the world of Harry Potter.
    \whp does not publish a full dataset.
    Hence, we use the snippets from the associated Hugging Face page,
    and we augment them with $20$ $(x_p\rightarrow c)$ Harry Potter trivia pairs generated with Llama2.
    \item \tofu~\cite{maini2024tofu} is a dataset of fictional writers, designed to be absent from the LLM's training data.
    Note that models released after \tofu was published might contain its records.
    We use the provided $10\%$ forget to $90\%$ retain split~\cite{tofuhf}.
\end{enumerate*}

\fauxsection{Models.}
We use Llama-2-7b-chat-hf~\cite{touvron2023llama} (Llama2), and Llama-3-8b-instruct~\cite{meta2024llama} (Llama3).
We get the unlearned \whp model from Hugging Face~\cite{eldan2023harrypotter-hf} (Llama2-\whp).

\fauxsection{Implementation.} We implement \sta using LLMart~\cite{llmart2025github} -- a PyTorch-based library for crafting adversarial prompts. 
We use implementations of the unlearning methods from the \tofu~\cite{tofugithub}, and NPO~\cite{npogithub} repositories.
We benchmark the attack against seven different unlearning algorithms: gradient ascent (GA), gradient difference (GDF)~\cite{liu2022continual}, refusal (IDK)~\cite{rafailov2024direct}, knowledge distillation (KL)~\cite{hinton2015distilling}, negative preference optimization (NPO)~\cite{zhang2024negative}, NPO-GDF, NPO-KL.

% We run our experiments on a machine equipped with Intel Xeon Gold 5218 CPU, eight NVIDIA A6000, and 256 GB of RAM.

\subsection{Auditing with attacks}\label{sec:eval:attacks}

\fauxsection{Who Is Harry Potter?.}
To elicit completions, we initialize the soft tokens using randomly selected hard tokens, and append them to the prompt $x_a = embed(x_p) \oplus embed(x_s)$.
We then train the soft prompt using AdamW~\cite{loshchilov2018decoupled} for up to $3000$ iterations; using $lr=0.005$, and $\beta{s}=(0.9, 0.999)$.
$x_p$ does not change, only the embedded suffix does.
If the optimization fails, we double the number of soft tokens up to the maximum of $16$.
We report the mean and standard deviation over five independent runs across all prompts.
In Table~\ref{tab:attack-unlearning-hp} we report the average number of soft tokens needed to elicit a completion. 
\whp* denotes the unlearned model with different prompt templates.

We show that all target completions can be generated with $\approx 4-6$ added soft tokens.
For all pairs of models, we conduct a \emph{t-test} under the null hypothesis $\mathcal{H}_0$ of \emph{equivalent population distributions} with $\alpha=0.05$.
We use an unpaired Welch's t-test since sample variances are not equal~\cite{welch1947ttest}.
We cannot reject the hypothesis for any of the pairs i.e. $p>0.05$.
In other words, for all models, no significant evidence that eliciting completions is more difficult.

Additionally, we observe that the ease of eliciting the completions changes depending on the prompt template.
We notice that the model (\whp+chat) reveals all unlearned information with manually paraphrased prompts (in a chat setting).
Furthermore, when using the example prompts in the corresponding Hugging Face repository~\cite{eldan2023harrypotter-hf}, Llama2-\whp would often begin the response with a double new line (\textbackslash n\textbackslash n).
We suspect that the provided unlearned model is overfit to ``\textit{prompt\textbackslash n\textbackslash n completion}''.
To run our evaluation in the most favorable setting, we report all three.
Our results show that attacking is the easiest for \textit{\whp+chat}, and the most difficult for \textit{\whp+\textbackslash n\textbackslash n}.
Given these discrepancies, and the lack of a standard \whp dataset, we believe it is not a good unlearning benchmark, despite its popularity.

% Also, in our dataset there are three challenging outlier prompts that require $16$ soft tokens, unlike other prompts.
% Filtering these out results in $4.05, 4.60, 1.61$ average required tokens for \whp, \whp+\textbackslash n\textbackslash n, and \whp+chat respectively.
\begin{table}[t]
    \centering
    \small
    \begin{subtable}[t]{\linewidth}
        \centering
        \begin{tabularx}{\linewidth}{|Y|Yc|}
            \hline
            Prompt                                & \multicolumn{2}{c|}{Model} \\
            template                              & Llama2-\whp    & Llama3 \\
            \hline
            N/A                                   & N/A            & $5.61\pm6.32$ \\
            \hline
            \whp                                  & $4.63\pm3.69$ & N/A \\
            \whp+\textbackslash n\textbackslash n & $6.50\pm5.13$ & N/A \\
            \whp+chat                             & $4.12\pm5.53$ & N/A \\
            \hline
        \end{tabularx}
        \caption{\whp results with different prompt templates.}
        \label{tab:attack-unlearning-hp}
    \end{subtable}
    \\
    \begin{subtable}[t]{\linewidth}
        \centering
        \begin{tabularx}{\linewidth}{|c|YY|}
            \hline
            \multirow{2}{*}{Unlearning method} & \multicolumn{2}{c|}{Model} \\
                                               & Llama2        & Llama3 \\
            \hline
            $f_\emptyset$ (none)               & $3.07\pm3.25$ & $3.11\pm3.15$ \\
            $f_{ft}$ (none)                    & $2.95\pm3.35$ & $3.21\pm3.19$ \\
            \hline
            $f_{u-IDK}$                        & $3.40\pm3.20$ & $3.33\pm3.09$ \\
            $f_{u-GA}$                         & $3.34\pm3.97$ & $3.21\pm3.87$ \\
            $f_{u-GDF}$                        & $3.06\pm3.34$ & $3.11\pm3.40$ \\
            $f_{u-KL}$                         & $3.08\pm3.31$ & $3.12\pm3.17$ \\
            $f_{u-NPO}$                        & $3.11\pm3.27$ & $3.12\pm3.27$ \\
            $f_{u-NPO-GDF}$                    & $3.15\pm3.24$ & $3.16\pm3.16$ \\
            $f_{u-NPO-KL}$                     & $3.23\pm3.62$ & $3.24\pm3.57$ \\
            \hline
        \end{tabularx} 
        \caption{\tofu results with different unlearning methods.}
        \label{tab:attack-unlearning-tofu}
    \end{subtable}
    \caption{Number of soft tokens needed to elicit a completion for a fixed number of iterations; averaged over all prompts in each set and over five runs per prompt.
    When increasing the maximum iterations to $10{,}000$, \textbf{all} completions can be elicited with $1\text{--}2$ soft tokens.}
    \vspace{-12pt}
    \label{tab:attack-unlearning-combined-subtables}
\end{table}

\fauxsection{TOFU.}
We follow the same setup as for \whp.
In Table~\ref{tab:attack-unlearning-tofu}, we report the number of soft tokens required to elicit the completions.
$f_\emptyset$ refers to the unmodified baseline model, $f_{ft}$ are the models fine-tuned on \tofu, followed by the unlearned models.

For all methods, we can elicit the completions with $\approx 3$ soft tokens.
Similarly to \whp, for all pairs of models (within the same architecture), we conduct Welch's \emph{t-test}.
We cannot reject the hypothesis for any of the pairs i.e. $p>0.05$; $f_{u-IDK}$ vs $f_{ft}$ (for Llama2) gives the lowest p-value of $0.509$.
For all models, there is not enough evidence to say that eliciting completions is more difficult.

One could argue that the unlearning methods used are not effective (when comparing $f_{ft}$ vs $f_{u-*}$), hence they require similar numbers of soft tokens.
In fact, most of these approaches have already been shown to be ineffective and susceptible to simple paraphrasing~\cite{shi2024muse}.
However, the same holds when compared to $f_\emptyset$.
In the next section, we demonstrate that the result cannot be attributed to the (in-)effectiveness of the unlearning methods but rather the power of \sta.

\subsection{Eliciting random strings}\label{sec:eval:random-strings}

The chance of a random string appearing in the training set is negligible, and preceding tokens do not inform the selection of the next token. 
We construct random strings uniformly at random from the range 33-126 of the ASCII table.

We initialize the soft prompt using randomly selected tokens.
In this experiment, there is only $x_s$, and no $x_p$.
We then train the soft prompt using AdamW for up to $3000$ iterations per soft token; using $lr=0.005$, and $\beta{s}=(0.9, 0.999)$.

\begin{figure}[t]
\resizebox{.85\columnwidth}{!}{\includegraphics{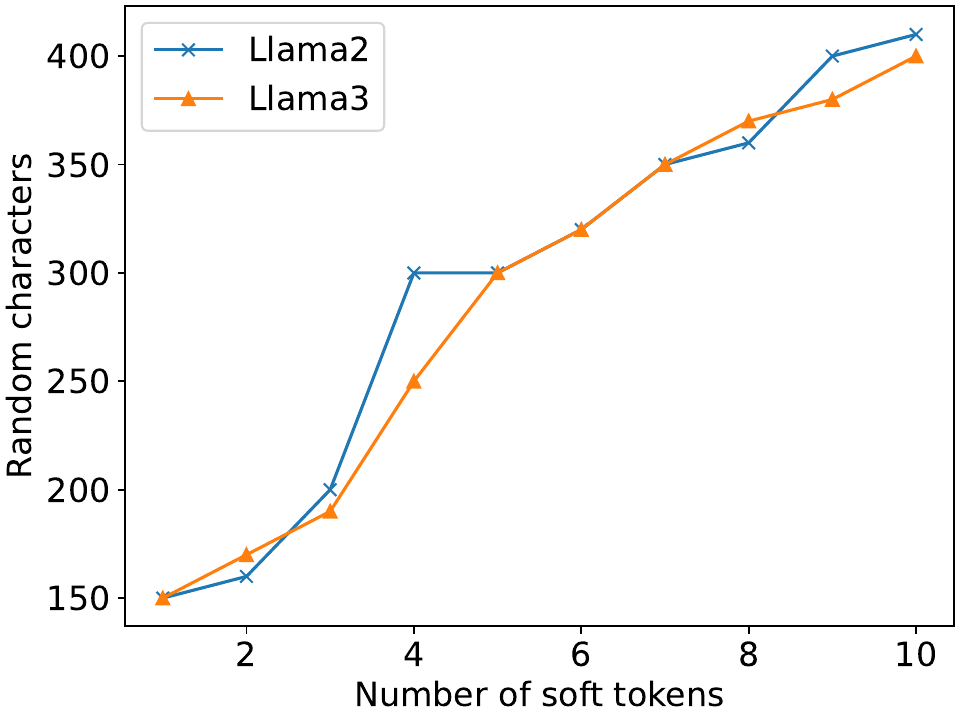}}
    \centering
    \vspace{-6pt}
    \caption{A single soft token can force over $150$ random characters. With $10$ soft tokens, it is possible to generate over $400$ random characters.}
    \vspace{-12pt}
    \label{fig:random_strings}
\end{figure}

In Figure~\ref{fig:random_strings}, we report the longest elicited string for a given number of soft tokens.
We repeat the experiment five times -- e.g., the first marker implies that for each of the five tested random strings of length $150$, we found an effective soft prompt.
We observe that not all initializations and seed configurations succeed, in which case a run needs to be restarted with a different seed.
If the loss plateaus around $25\%$ of the iterations, we restart the run.
However, no single string was restarted more than ten times.
Our results show that \sta{s} can be used to elicit completely random strings, thus undermining their application for auditing unlearning.

Next, we aim to answer why eliciting strings is possible.
Prompt-tuning~\cite{lester2021prompttuning} is an efficient fine-tuning technique that trains only a soft prompt added to the input instead of all weights.
\sta{s} can be viewed as an extreme case of prompt-tuning, where instead of training over many prompts, one trains an attack per each prompt.
Thus, an LLM that outputs a completion that it was trained on is an expected behavior. 
However, one could argue, in practice, a properly unlearned (or aligned) LLM should never output undesirable text.

%% file: 5discussion.tex
\section{Discussion \& Conclusion}\label{sec:discussion}

% \fauxsection{Auditing with hard prompts.}
% Attacks like greedy coordinate gradient~\cite{zou2023gcg} optimize the attack prompt in the \emph{hard} token space instead of the soft token space.
% Hence, they are weaker at eliciting completions.
% While this might make them more suitable for auditing unlearning, due to their computational requirements, they are often used to force only the beginning of a harmful completion (e.g. \textit{Sure, here's how to build a bomb...}) with the hope that the LLM follows.
% It is unclear whether this would be sufficient to produce specific unlearned passages.
% We see it as an interesting direction for future work.
\fauxsection{Unlearning vs jailbreaking.}
Our findings also apply to the jailbreaking community.
Prior work hinted that unlearning and preventing harmful outputs can be both viewed as suppressing particular information~\cite{zhang2024safe}.
For instance, it was shown that fine-tuning on benign or unrelated data can restore undesirable behavior~\cite{hu2024jogging,lucki2024adversarial}.

\fauxsection{Variation in gradient descent.}
Prior work showed that retraining with some records removed can result in the same final model depending on the seed~\cite{thudi2022auditunlearning}.
The influence of the records might be minimal, making unlearning unnecessary.
Similarly, it was shown that SGD has intrinsic privacy guarantees, assuming there exists a group of similar records~\cite{hyland2022empiricalsgdprivacy}.
Thus, algorithmic auditing of unlearning might not be possible, and one would have to rely on verified or attested procedures instead~\cite{eisenhofer2023verifiedunlearning}, regardless of their impact on the model.

\fauxsection{Distinguishing learned soft tokens.}
Even though, in most of our results, the number of soft tokens required to elicit a completion is the same,
we attempt to distinguish between them.
To this end, we take all single-token \sta{s} optimized for \tofu (Table~\ref{tab:attack-unlearning-tofu}) and assign a label $y=\{0, 1\}$: $y=0$ for $f_\emptyset$, and $y=1$ for $f_{ft}$ and the unlearned models.
We then train a binary classifier using $f_\emptyset$ and $f_{ft}$.
While we are able to overfit it and distinguish between $f_\emptyset$ and $f_{ft}$,
we were not able to train a model that would generalize to the unlearned models, and decisively assign a class. 
Our approach is similar to Dataset Inference~\cite{maini2021di,maini2024dillms} which showed there can be distributional differences between the models, depending on the data they were trained on.
Further investigation into \emph{what} soft tokens are learned during the audit is an interesting direction for future work.

\fauxsection{Conclusion.}\label{sec:conclusion}
In this work, we show that soft token attacks (\sta) cannot distinguish between base, fine-tuned, and unlearned models: a strong auditor can elicit all unlearned text.
Also, we show that \sta with a single token can elicit $150$ random characters, and over $400$ with $10$ tokens.
Our work shows that machine unlearning in LLMs needs careful assumptions to avoid misleading results.

\section{Limitations \& ethical considerations}\label{sec:limitations}

\fauxsection{Limitations.}
Our experiments are constrained to models with 7–8 billion parameters. 
Nevertheless, since the expressive power of LLMs generally scales with size~\cite{kaplan2020scalinglaws}, we expect our findings to extend to larger models. 
For efficiency reasons, we restrict our evaluation to at most $10$ soft tokens and random strings of up to $400$ characters, which does not establish an upper bound on the length of strings that can be elicited. 
Future work could investigate whether black-box optimization methods—such as zeroth-order optimization~\cite{chen2017zoo}—can reproduce the elicitation observed in the white-box setting. 
In addition, extending our evaluation with random strings may help determine whether there exists a clear and generalizable relationship between the number of soft tokens and the maximum number of generated characters.

\fauxsection{Ethical considerations.}
In this work, we show that an auditor with white-box access and sufficient computational resources can elicit arbitrary text from an LLM. Although this requires knowledge of the target completion, partial completions may suffice, enabling the extraction of harmful information—especially when the auditor has approximate prior knowledge of the removed content.

%% file: 6acknowledgments.tex
\section{Acknowledgments}

We thank Cory Cornelius and Marius Arvinte for valuable discussions on this work.

%% file: appendix.tex
\clearpage
\begin{appendices}

\begin{figure*}[t]
    \includegraphics[width=\textwidth]{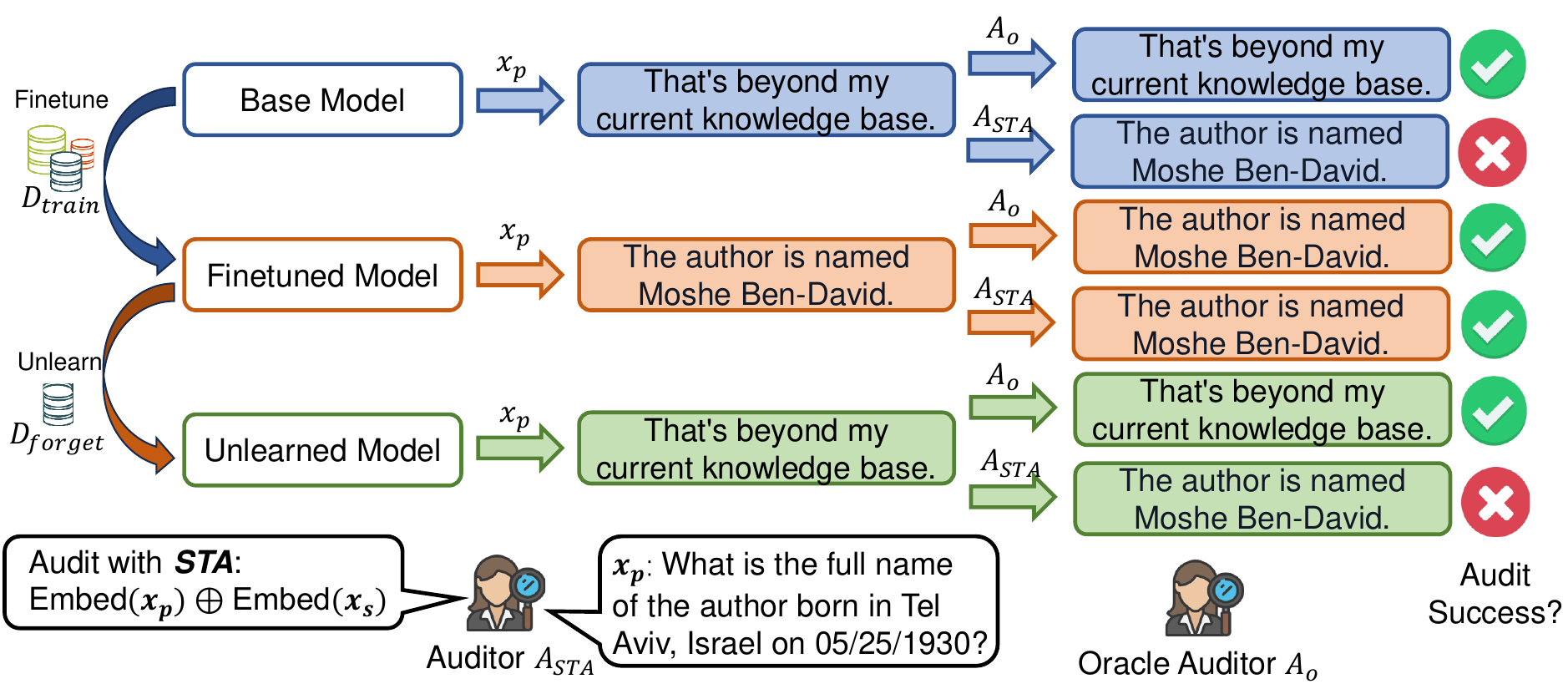}
    \caption{Overview of the auditing process using $A_\sta$. For a perfect unlearning method, $A_o$ always correctly audits the model. On the other hand, $A_\sta$ can elicit the completion regardless of the information in the model -- the audit is ineffective.}
    \label{fig:auditor}
\end{figure*}

\begin{table}[t]
    \centering
    \begin{tabular}{|c|c|}
        \hline
        \sta          & soft token attack\\
        $A_o$         & oracle auditor\\
        $A_\sta$      & \sta auditor\\
        $x_p$         & base prompt (benign)\\
        $x_s$         & adversarial suffix\\
        $x_a$         & adversarial prompt ($x_p \oplus x_s$)\\
        $c$           & target completion\\
        $f_\emptyset$ & base model\\
        $f_{ft}$      & fine-tuned model\\
        $f_{u}$       & unlearned model\\
        $f_{u-*}$     & model unlearned using *\\
        $D_{train}$   & training data\\
        $D_{forget}$  & forget data\\
        $D_{retain}$  & retain data\\
        \hline
    \end{tabular}
    \caption{Summary of the notation. '*' is replaced with the specific unlearning method.}
    \label{tab:notation}
\end{table}

% \begin{algorithm}[h!]
% \caption{\textsc{STA: Soft Token Auditor}}
% \begin{algorithmic}[1]
% \State \textbf{Input:} Finetuned model $f_{\text{ft}}$, unlearned model $f_u$, unlearned query $x_q$
% \State \textbf{Output:} Score $s$ indicating unlearning effectiveness

% \vspace{0.5em}
% \Function{FindMinTokenLength}{$f, x_p, x_s$}
%     \For{$k = 1$ \textbf{to} $5$}
%         \State Randomly initialize $2^k$ soft token embeddings $e$
%         \State Optimize $e$ so that $f(\text{embed}(x_p) \oplus e)$ generates target output $c$
%         \If{success}
%             \State \Return $2^k$
%         \EndIf
%     \EndFor
%     \State \Return $+\infty$ \Comment{Attack failed within budget}
% \EndFunction

% \vspace{0.5em}
% \State $k_u \gets$ \Call{FindMinTokenLength}{$f_u, x_p, x_s$}
% \State $k_{\text{ft}} \gets$ \Call{FindMinTokenLength}{$f_{\text{ft}}, x_p, x_s$}
% \State \Return $s \gets k_u - k_{\text{ft}}$

% \end{algorithmic}
% \end{algorithm}

\section{Auditing process}\label{appendix:process}
Figure~\ref{fig:auditor} gives a complete overview of the auditing procedure, and the difference between $A_o$ and $A_\sta$.
In Table~\ref{tab:notation}, we summarize the notation.

\end{appendices}